\documentclass[lettersize,journal]{IEEEtran}
\usepackage{subcaption}
\usepackage{caption}
\usepackage{booktabs}
\usepackage[T1]{fontenc}
\usepackage{amsmath,amsfonts}
\usepackage{algorithmic}
\usepackage{algorithm}
\usepackage{array}
\usepackage{textcomp}
\usepackage{stfloats}
\usepackage{url}
\usepackage{verbatim}
\usepackage{graphicx}
\usepackage{cite}

\usepackage{multirow}
\begin{document}

\title{SPGCL: Simple yet Powerful Graph Contrastive Learning via SVD-Guided Structural Perturbation}

\author{Hao Deng, Zhang Guo,~\IEEEmembership{Member,~IEEE}, Shuiping Gou, ~\IEEEmembership{Senior Member,~IEEE}, and Bo Liu, ~\IEEEmembership{Member,~IEEE}
\thanks{This work was jointly supported by the National Natural Science Foundation of China NSFC under Grant T2541002, Grant 62372358 and Grant 62302355.(Corresponding authors: Zhang Guo; Bo Liu)}
\thanks{Hao Deng, Zhang Guo, Shuiping Gou and Bo Liu are with School of Artificial Intelligence, Xidian University, Xi'an, Shaanxi, 710126, China(emails: dengkoicat@yeah.net, guozhang@xidian.edu.cn
, shpgou@mail.xidian.edu.cn, liub@xidian.edu.cn)}
}

\maketitle

\begin{abstract}
Graph Neural Networks (GNNs) are highly sensitive to structural noise, including spurious or missing edges caused by adversarial attacks or non-adversarial imperfections. Existing graph contrastive learning methods typically rely on either random perturbations (e.g., edge dropping) to ensure view diversity, or purely spectral augmentations (e.g., SVD) to preserve global structural priors. However, random perturbations are structure-agnostic and may inadvertently remove critical edges, while purely SVD-based views tend to be dense and lack sufficient diversity. \textbf{More fundamentally, despite their complementary strengths, there exists no straightforward way to integrate these two fundamentally different paradigms: one operates on discrete edge removal while the other relies on continuous matrix factorization.}

To bridge this \textbf{non-trivial} gap, we propose SPGCL, a simple yet powerful framework for robust graph contrastive learning via SVD-guided structural perturbation. \textbf{Our key insight lies in leveraging a recently developed SVD-based perturbation method}, \textbf{which generalizes the classical structural perturbation theory from symmetric to arbitrary graphs.} We creatively adapt this technique to contrastive learning by designing a concise two-stage perturbation strategy: first, a lightweight stochastic edge removal injects necessary diversity; second, truncated SVD is applied to derive a structure-aware scoring matrix, followed by sparse top-$P$ edge recovery. This synergistic integration delivers three key advantages: (1) \textbf{Robustness to accidental deletion} - important edges removed in the first stage can be recovered by SVD-guided scoring, mitigating harmful effects while preserving diversity. (2) \textbf{Enrichment with missing links} - the SVD-based scoring can introduce new, semantically meaningful edges absent in the original graph, creating a more informative contrastive view. (3) \textbf{Controllable structural discrepancy} – by balancing edge removal and recovery rates, we explicitly control the edge count between views, ensuring that the contrastive signal stems from semantic structural differences rather than trivial edge-number gaps.

Furthermore, we incorporate a contrastive fusion module regularized by a global similarity constraint to align node embeddings across views. Extensive experiments on ten benchmark datasets demonstrate that SPGCL consistently improves the robustness and accuracy of base GNNs, outperforming state-of-the-art graph contrastive learning and structure learning methods, thereby validating its simplicity, effectiveness, and \textbf{innovative integration of previously disparate paradigms}.
\end{abstract}


\section{Introduction}

Graph Neural Networks (GNNs) have become the de facto standard for learning representations from relational data, achieving remarkable success in domains ranging from social network analysis to biochemical discovery \cite{kipf2017semi, wu2019simplifying,hamilton2017inductive, velickovic2018graph, xu2019powerful}. Their core strength lies in recursively aggregating information from neighboring nodes, effectively leveraging both node attributes and graph topology.

However, this strength is also a critical vulnerability: GNNs are highly sensitive to imperfections in the input graph structure \cite{zugner2018adversarial}. In practice, graphs are often noisy due to adversarial attacks \cite{zugner2018adversarial,pmlr-v80-dai18b,wu2019adversarial} (e.g., fake links in recommendation systems) \cite{ying2018graph,wang2019neural} or non-adversarial imperfections (e.g., missing observations or measurement errors). Such structural noise can corrupt the message-passing process, leading to severe degradation in model performance and robustness.

To mitigate this issue, Graph Contrastive Learning (GCL) has emerged as a powerful self-supervised paradigm. By contrasting different “views” of the same graph, GCL encourages the model to learn representations that are invariant to benign perturbations \cite{oord2018representation, youGraphContrastiveLearning2020}. The crux of GCL lies in designing effective graph augmentation strategies to generate these views. Prevailing methods largely fall into two categories. The first employs random perturbations, such as uniformly dropping edges \cite{zhu2020deep}. While simple and conducive to view diversity, these operations are structure-agnostic and may inadvertently remove semantically critical edges, providing noisy or even detrimental contrastive signals. The second category leverages spectral methods, most notably using Singular Value Decomposition (SVD) to generate a globally informed view \cite{cai2023lightgcl}. This approach preserves high-order structural patterns but often results in a dense and relatively deterministic view, which may limit the diversity crucial for contrastive learning and introduce many weak connections.

\textbf{The Fundamental Challenge of Integration.}
While the complementary strengths of random and spectral augmentations are intuitively appealing, their integration poses a \emph{non-trivial} challenge. The two paradigms operate on fundamentally different principles: random edge removal is a \emph{discrete, structure-agnostic} operation, whereas SVD-based augmentation is a \emph{continuous, global} matrix factorization process. There exists no natural bridge between these two distinct mechanisms, making their synergistic combination far from straightforward. Simply applying both in sequence (e.g., SVD followed by random dropping) often leads to conflicting effects-the global consistency captured by SVD can be arbitrarily destroyed by subsequent random removal. Therefore, despite the clear potential benefits, how to \emph{seamlessly and effectively integrate} these disparate approaches remains an open and under-explored problem.

\textbf{Our Insight: SVD-Guided Perturbation as the Bridge.}
To overcome this challenge, we draw inspiration from a recently developed \emph{SVD-based perturbation method} \cite{deng2026gadpn}, which generalizes the classical structural perturbation theory \cite{lu2015toward} from symmetric to arbitrary graphs via singular value decomposition. This method provides a principled way to quantify the structural impact of edge modifications (addition/removal) through first-order singular value shifts, effectively establishing a continuous link between discrete edge operations and the spectral domain. We recognize that this technique can serve as the missing link to bridge random and spectral augmentations: the stochastic removal step provides the necessary diversity, while the SVD-based perturbation theory enables a \emph{structure-aware recovery and enrichment} of edges, ensuring global consistency is preserved.

\textbf{Proposed Framework: SPGCL.}
Building on this insight, we propose \textbf{SPGCL}, a \emph{simple yet powerful} framework for robust graph contrastive learning via SVD-guided structural perturbation. Our core innovation is a \emph{coupled two-stage augmentation strategy}: a lightweight stochastic edge removal followed by an SVD-guided scoring and sparse edge recovery. This design is not merely a combination but a synergistic integration that delivers distinct advantages:

\begin{itemize}
\item \textbf{Risk Mitigation for Random Deletion}: If the initial random drop accidentally removes an important edge, our SVD-guided recovery step can, with high probability, restore it. This preserves the beneficial diversity introduced by randomness while dramatically reducing its potential harm.
\item \textbf{Enrichment Beyond the Original Graph}: Beyond recovery, the SVD-based scoring can identify and add new, semantically meaningful edges that are missing from the original graph, thereby creating a more informative contrastive view.
\item \textbf{Controlled Structural Discrepancy}: By balancing the rates of edge removal and recovery, we can explicitly control the total edge count in the perturbed view. This ensures the contrastive signal stems from meaningful structural semantics rather than trivial differences in edge density.
\end{itemize}

Furthermore, we introduce a contrastive fusion module with a global similarity-matrix constraint to align node embeddings from the original and perturbed views, ensuring consistency in both local and global structural patterns.

Extensive experiments on ten benchmark datasets demonstrate that SPGCL consistently outperforms state-of-the-art GCL and graph structure learning baselines. Ablation studies confirm the contribution of each component, and robustness analyses show our model's superior stability under increasing structural noise.

\textbf{Our main contributions are summarized as follows:}

\begin{itemize}
\item We identify and articulate the fundamental challenge of integrating random and spectral augmentations in graph contrastive learning, highlighting the disparate nature of the two paradigms.
\item We introduce a novel \textbf{SVD-guided structural perturbation} strategy that creatively adapts a recent SVD-based perturbation method \cite{deng2026gadpn} to bridge random diversity with spectral rationality, enabling their synergistic integration for the first time.
\item Our design delivers three key benefits: \textbf{robustness to accidental edge deletion}, \textbf{enrichment via missing link addition}, and \textbf{controlled structural discrepancy}, advancing the design of graph augmentations.
\item We incorporate a \textbf{contrastive fusion scheme} with a global consistency constraint to effectively align multi-view representations.
\item We provide comprehensive empirical evidence of SPGCL's effectiveness and robustness across diverse real-world graphs.
\end{itemize}

\section{Related Work}

\subsection{Robustness in Graph Neural Networks}
The sensitivity of GNNs to structural noise has spurred extensive research into robust learning. Early adversarial defense works focus on identifying and hardening models against malicious perturbations \cite{zugner2018adversarial,wu2019adversarial,entezari2020all,tang2020transferring,xu2022unsupervised,zheng2020robust}. Another line of work, \textbf{Graph Structure Learning (GSL)}, aims to infer an optimal or denoised graph topology jointly with GNN training. Methods like LDS \cite{franceschi2019learning} and Pro-GNN \cite{jin2020graph} impose constraints (e.g., low-rank, sparsity) to reconstruct cleaner graphs. While improving robustness, these methods often operate in a \textbf{single-view} manner and may not explicitly preserve the diverse, task-agnostic signals crucial for generalizable representation learning.

\subsection{Graph Contrastive Learning and Augmentation Strategies}
\textbf{Graph Contrastive Learning (GCL)} provides a self-supervised pathway to robust representations by contrasting multiple augmented views of the graph \cite{youGraphContrastiveLearning2020, zhu2020deep}. The performance of GCL hinges critically on the design of \textbf{graph augmentation}. Existing augmentation strategies can be categorized along a spectrum from \textbf{local/random} to \textbf{global/spectral}:

\begin{itemize}
\item \textbf{Local Random Perturbations:} Most GCL frameworks \cite{zhu2020deep, zhu2021graph} rely on random operations like edge dropping and feature masking. These are simple and promote view diversity. However, their \emph{structure-agnostic} nature is a fundamental limitation: they risk removing important edges and fail to inject semantically meaningful structural information, making the contrastive signal potentially noisy or uninformative.
\item \textbf{Global Spectral Augmentations:} To capture higher-order and global structural patterns, recent methods like LightGCL \cite{cai2023lightgcl} employ Singular Value Decomposition (SVD) to generate augmented views. While more structure-aware, purely SVD-based views tend to become \emph{dense and deterministic}, which can reduce the effective diversity between views and introduce many weak connections.
\end{itemize}

This dichotomy highlights a \textbf{critical gap}: existing methods either prioritize \emph{diversity at the cost of structural rationality} (random) or prioritize \emph{global consistency at the cost of diversity and sparsity} (spectral). A principled integration that harnesses the benefits of both paradigms remains underexplored.

\subsection{Structural Perturbation and Its SVD-based Generalization}
\label{subsec:structural_perturbation}
\textbf{Structural Perturbation Method (SPM)} pioneered by Lü et al.~\cite{lu2015toward} provides a principled framework for link prediction by analyzing the stability of a network's eigenvectors under small perturbations. Rooted in first-order eigenvalue perturbation theory, SPM offers a global, structure-aware perspective on link predictability, moving beyond local heuristics. However, it is inherently limited to \emph{symmetric, undirected graphs} due to its reliance on eigendecomposition, restricting its direct application to general graph types.

To overcome this limitation, \textbf{GADPN}~\cite{deng2026gadpn} generalizes SPM to arbitrary (including directed and attributed) graphs via \emph{Singular Value Decomposition (SVD)}. By deriving a first-order singular value perturbation formula $\Delta \sigma_i \approx \mathbf{u}_i^\top \Delta \mathbf{A} \mathbf{v}_i$, GADPN extends the original eigenvalue-based perturbation theory to general adjacency matrices. This extension not only enables structure-aware edge recovery and denoising in broader graph learning scenarios but also establishes a \emph{continuous link} between discrete edge modifications and the spectral domain-a crucial insight that bridges operations on graph structures with their low-rank approximations.

While GADPN focuses on \emph{graph structure learning} for semi-supervised node classification, its core SVD-based perturbation technique provides a \emph{generic and powerful tool} to quantify and compensate for structural changes. In this work, we recognize that this very technique can be creatively repurposed to address a fundamentally different challenge: \emph{integrating random and spectral augmentations in graph contrastive learning}. SPGCL thus stands as an innovative application of the generalized perturbation framework, demonstrating its versatility and potential to connect previously disparate paradigms in graph representation learning.

\subsection{Beyond Diversity: Towards Rational and Controlled Perturbation}
Recent studies have begun to scrutinize the quality of augmentations. Some works point out the instability of random augmentations \cite{yuAreGraphAugmentations2022}, while others attempt to guide perturbations with node features or gradients \cite{zhu2021graph}. However, these approaches often lack a \textbf{global structural prior} or do not address the \textbf{post-perturbation recovery} of critical edges. Crucially, few methods consider \textbf{explicitly controlling the structural discrepancy} between views (e.g., edge count), leaving open the risk that contrastive learning might exploit trivial, task-irrelevant differences.

Our work directly addresses these shortcomings. SPGCL introduces a novel \textbf{SVD-guided structural perturbation} that not only \emph{couples} random diversity with spectral rationality but also provides a \textbf{recovery mechanism} for accidentally deleted edges, \textbf{enriches} the view with missing high-order links, and enables \textbf{explicit control} over the perturbation scale. By building upon the generalized perturbation theory reviewed in Section~\ref{subsec:structural_perturbation}, our method offers a principled and interpretable approach to graph augmentation, positioning SPGCL as a step towards more effective and robust contrastive learning on graphs.

\section{SVD-Guided Structural Perturbation}

This section details the core component of SPGCL: an SVD-guided structural perturbation strategy designed to generate an augmented view that is both diverse and structure-aware. We first introduce the necessary preliminaries, then elaborate on the motivation and the step-by-step perturbation process.

\subsection{Preliminaries}
\subsubsection{Problem Setting}
We consider an attributed graph $\mathcal{G} = (\mathcal{V}, \mathcal{E}, \mathbf{X})$,
where $\mathcal{V} = \{v_1, v_2, \dots, v_N\}$ is the node set with $|\mathcal{V}| = N$,
$\mathcal{E} \subseteq \mathcal{V} \times \mathcal{V}$ is the edge set,
and $\mathbf{X} \in \mathbb{R}^{N \times d}$ is the node feature matrix.
The graph structure is represented by the adjacency matrix $\mathbf{A} \in \{0,1\}^{N \times N}$,
where $\mathbf{A}_{ij} = 1$ if $(v_i, v_j) \in \mathcal{E}$ and $\mathbf{A}_{ij} = 0$ otherwise.
We focus on semi-supervised node classification, where only a subset of nodes $\mathcal{V}_L \subseteq \mathcal{V}$ are labeled.

\subsubsection{Graph Convolutional Network Backbone}
We adopt the Graph Convolutional Network (GCN) \cite{kipf2017semi} as our encoder. Each layer propagates information as:
\begin{equation}
\mathbf{H}^{(l+1)} = \sigma \left( \tilde{\mathbf{D}}^{-\frac{1}{2}} \tilde{\mathbf{A}} \tilde{\mathbf{D}}^{-\frac{1}{2}} \mathbf{H}^{(l)} \mathbf{W}^{(l)} \right),
\end{equation}
where $\tilde{\mathbf{A}} = \mathbf{A} + \mathbf{I}$, $\tilde{\mathbf{D}}$ is the diagonal degree matrix of $\tilde{\mathbf{A}}$, and $\sigma(\cdot)$ is a nonlinear activation.

\subsubsection{Low-Rank Approximation via SVD}
Singular Value Decomposition (SVD) provides a low-rank approximation of the adjacency matrix:
\begin{equation}
\mathbf{A} = \mathbf{U} \boldsymbol{\Sigma} \mathbf{V}^{\mathsf{T}} = \sum_{i=1}^{N} \sigma_i \mathbf{u}_i \mathbf{v}_i^{\mathsf{T}}, 
\end{equation}
where $\mathbf{U}, \mathbf{V}$ are orthogonal matrices, $\boldsymbol{\Sigma} = \text{diag}(\sigma_1, \dots, \sigma_N)$ with $\sigma_1 \geq \dots \geq \sigma_N \geq 0$. A rank-$r$ truncated SVD retains the top-$r$ components to capture the most significant global structural patterns:
\begin{equation}
\mathbf{A} \approx \sum_{i=1}^{r} \sigma_i \mathbf{u}_i \mathbf{v}_i^{\mathsf{T}}.
\end{equation}

\subsection{Motivation and Design Intuition}
\label{subsec:Motivation and Design Intuition}
Existing augmentation strategies for graph contrastive learning suffer from a fundamental trade-off: random perturbations (e.g., edge dropping) are diverse but structure-agnostic, while purely spectral augmentations (e.g., SVD) are structure-aware but tend to be dense and deterministic. To overcome this trade-off and achieve robust representations, our goal is to generate a perturbed view that simultaneously fulfills three objectives:
\begin{enumerate}
\item \textbf{Preserving diversity} for effective contrastive learning,
\item \textbf{Maintaining global structural consistency} by avoiding arbitrary corruption of important edges,
\item \textbf{Ensuring sparsity} to avoid introducing excessive weak connections.
\end{enumerate}

These objectives directly motivate the three key advantages of our method: robustness to accidental deletion, enrichment with missing links, and controllable structural discrepancy.

To this end, we propose a two-stage structured perturbation that couples stochastic edge removal with SVD-guided refinement. This simple yet principled design not only balances diversity and structural awareness but also naturally delivers the aforementioned advantages:

\begin{itemize}
\item \textbf{Robust recovery of accidentally removed edges}: Important edges dropped in the first stage can be recovered via SVD-guided scoring.
\item \textbf{Enrichment with missing high-order connections}: The SVD-based scoring can identify and add semantically meaningful edges absent in the original graph.
\item \textbf{Controllable structural discrepancy}: By balancing removal and addition, we explicitly control the edge count, ensuring the contrastive signal stems from semantic differences rather than mere edge-number gaps.
\end{itemize}

\subsection{Two-Stage Structured Perturbation}
\label{sec:sv_perturb}

\begin{figure*}[htbp]
  \centering
  \includegraphics[width=0.8\textwidth]{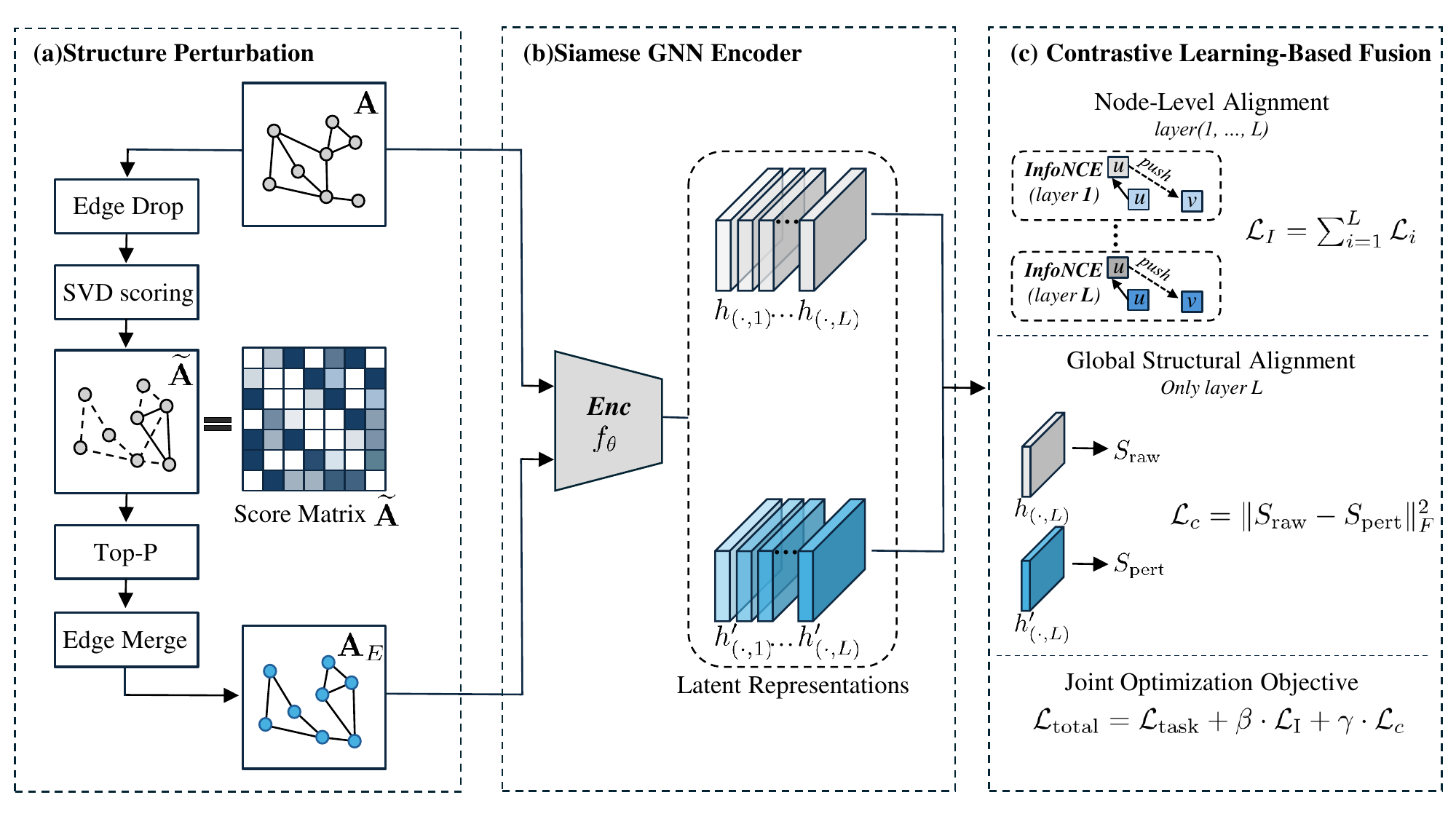}

  \caption{Overview of the proposed SPGCL framework.
  \textbf{(a) SVD-guided structured perturbation.} Starting from the original adjacency $\mathbf{A}$, we first apply \emph{Edge Drop} to obtain the remaining graph $\mathbf{A}_R$ and the removed edges $\Delta \mathbf{A}$.
  We then perform truncated SVD on $\mathbf{A}_R$ and estimate the singular-value perturbation via a first-order approximation, constructing a topology-aware \emph{score matrix} $\widetilde{\mathbf{A}}$.
  To avoid graph densi\-fication, we select the \emph{Top-$P$} highest-scored candidate edges (excluding edges in $\mathcal{E}_R$) to form a sparse addition matrix $\mathbf{A}_P$, and merge it with $\mathbf{A}_R$ to produce the sparse perturbed view $\mathbf{A}_E = \mathbf{A}_R + \alpha \mathbf{A}_P$ Eq~\eqref{eq:a_e}.
  \textbf{(b) Siamese GNN encoder.} A shared encoder $f_\theta$ extracts layer-wise node representations from the original view ($\mathbf{A}$) and the perturbed view ($\mathbf{A}_E$).
  \textbf{(c) Contrastive learning-based fusion.} We apply an InfoNCE-based alignment loss $\mathcal{L}_I$ at \emph{each} encoder layer for multi-level node-wise agreement, while enforcing a global similarity-matrix consistency loss $\mathcal{L}_c$ \emph{only} on the final layer to regularize global structural patterns.
  All components are jointly optimized with the task loss.}

  \label{fig:framework}
\end{figure*}

Our perturbation process, illustrated in Fig.~\ref{fig:framework}, consists of two synergistic stages.

\paragraph{Stage 1: Lightweight Stochastic Edge Removal}
We first randomly remove a small fraction $p$ of edges from $\mathcal{E}$ to form a removed set $\Delta\mathcal{E}$. The remaining edge set is $\mathcal{E}_R = \mathcal{E} \setminus \Delta\mathcal{E}$, with corresponding adjacency matrices $\Delta\mathbf{A}$ and $\mathbf{A}_R$, such that $\mathbf{A} = \mathbf{A}_R + \Delta\mathbf{A}$. This step injects necessary diversity into the view.

\paragraph{Stage 2: SVD-Guided Scoring and Sparse Recovery}
We then perform truncated SVD on $\mathbf{A}_R$:
\begin{equation}
\mathbf{A}_R \approx \mathbf{U}_r \boldsymbol{\Sigma}_r \mathbf{V}_{r}^{\mathsf{T}}
= \sum_{i=1}^{r} \sigma_i \mathbf{u}_i \mathbf{v}_i^{\mathsf{T}} .
\end{equation}
In practice, we adopt the \emph{randomized SVD} algorithm \cite{halko2011finding} to efficiently compute the rank-$r$ approximation. 

\textbf{Key step: estimating structural impact via generalized perturbation.}
To account for the structural impact of the removed edges in a principled manner, we leverage the \emph{generalized singular value perturbation theory} recently developed in \cite{deng2026gadpn}. This theory extends the classic eigenvalue perturbation method \cite{lu2015toward} from symmetric to arbitrary graphs via SVD, providing a first-order approximation of how edge modifications affect the singular values:
\begin{equation}
\Delta \sigma_i \approx \mathbf{u}_i^{\mathsf{T}} \Delta\mathbf{A} \mathbf{v}_{i}.
\label{eq:singular_perturbation}
\end{equation}
Eq.~\eqref{eq:singular_perturbation} is crucial: it establishes a continuous, differentiable relationship between discrete edge removal ($\Delta\mathbf{A}$) and the spectral domain ($\mathbf{u}_i, \sigma_i, \mathbf{v}_i$). This exactly serves as the missing bridge that allows us to connect the stochastic removal in Stage 1 with the spectral recovery in Stage 2.

A refined score matrix is then constructed by adjusting the singular values according to their estimated shifts:
\begin{equation}
\widetilde{\mathbf{A}} = \sum_{i=1}^{r} (\sigma_i + \Delta \sigma_i)\mathbf{u}_i \mathbf{v}_i^{\mathsf{T}} .
\end{equation}
This matrix encodes global structural information while explicitly reflecting the perturbation introduced by the removed edges.

\textbf{Top-$P$ sparsification.}
Importantly, we sparsify $\widetilde{\mathbf{A}}$ to avoid densification \cite{cai2023lightgcl}.
We rank candidate edges not in $\mathcal{E}_R$ by their scores in $\widetilde{\mathbf{A}}$, select the top-$P$ entries,
and form a sparse addition matrix $\mathbf{A}_P$.
The number of recovered edges $P$ is determined by:
\begin{equation}
P=\max\!\left(\lfloor (p+q)|\mathcal{E}|\rfloor,0\right),
\label{eq:P_pq}
\end{equation}
where $p$ is the edge drop ratio, and it also serves as the base scale for edge recovery
and $q$ is a flexible adjustment parameter \cite{deng2026gadpn}.
In practice, $q$ can be positive, zero, or negative, thereby controlling whether we recover more, the same, or fewer edges than those initially removed.

The final perturbed adjacency is:
\begin{equation}
\mathbf{A}_E = \mathbf{A}_R + \alpha \cdot \mathbf{A}_P,
\label{eq:a_e}
\end{equation}
where $\alpha \in (0,1]$ controls the strength (weight) of the added edges. This yields a sparse, structure-aware augmented view ready for contrastive learning.

\paragraph{Summary of Advantages.}
Our two-stage perturbation directly addresses the objectives outlined in Sec.~\ref{subsec:Motivation and Design Intuition} and embodies the \emph{simple yet powerful} design philosophy:
\begin{itemize}
\item The random removal ensures \textbf{diversity} in a lightweight manner.
\item The SVD-guided scoring, built upon the generalized perturbation theory \cite{deng2026gadpn}, preserves \textbf{global structural patterns} and enables \textbf{recovery} of important edges by mathematically linking edge-level changes to spectral shifts.
\item The top-$P$ sparsification maintains \textbf{sparsity} and provides \textbf{explicit control} over the edge count and structural discrepancy.
\end{itemize}
This principled and modular design, grounded in a rigorous perturbation framework, forms the foundation of SPGCL, enabling robust and informative contrastive learning even under structural noise with minimal complexity overhead.

\section{The Proposed Method: SPGCL}
\label{sec:method}

In this section, we present the overall architecture of SPGCL, a \emph{simple yet powerful} framework for robust graph contrastive learning. It consists of two seamlessly integrated components: (1) an \textbf{SVD-guided structural perturbation} module that generates a sparse, structure-aware augmented view, and (2) a \textbf{contrastive learning-based fusion} module that aligns the original and perturbed views while enforcing global structural consistency. The complete pipeline is illustrated in Fig.~\ref{fig:framework}.

\subsection{Framework Overview}

SPGCL follows a dual-view contrastive learning paradigm. Given an input graph $\mathcal{G} = (\mathcal{V}, \mathcal{E}, \mathbf{X})$ with adjacency matrix $\mathbf{A}$, we first generate a perturbed view $\mathbf{A}_{E}$ via the structured perturbation described in Section~III. Both the original graph (with adjacency $\mathbf{A}$) and the perturbed graph (with adjacency $\mathbf{A}_{E}$) are then encoded by a shared GNN encoder $f_{\theta}$ to obtain node embeddings. Finally, a contrastive fusion module aligns the two views at both node and global structural levels.

\subsection{Contrastive Learning-Based Fusion}
\label{sec:fusion}

While the SVD-guided perturbation generates a structurally informed view, the inherent discrepancy between the original and perturbed views may lead to inconsistent node representations if not properly aligned. To address this, we introduce a contrastive fusion scheme that operates at two levels: node-level alignment via InfoNCE loss and global structural alignment via a consistency constraint.

\paragraph{Node-Level Alignment via InfoNCE}
We employ a Siamese GNN encoder $f_{\theta}$ to extract layer-wise node embeddings from both views. Let $h_{(u,l)}$ and $h'_{(u,l)}$ denote the embeddings of node $u$ at the $l$-th layer from the perturbed and original views, respectively. Following \cite{oord2018representation}, we apply the InfoNCE loss at each layer to maximize the agreement between corresponding node embeddings across views:
\begin{equation}
\mathcal{L}_{\text{I}} = -\frac{1}{N} \sum_{u=1}^{N}\sum_{l=1}^{L}\log \frac{\exp\left(\text{sim}\left(h_{(u,l)}, h'_{(u,l)}\right) / \tau\right)}{\sum_{v=1}^{N} \exp\left(\text{sim}\left(h_{(u,l)}, h'_{(v,l)}\right) / \tau\right)},
\end{equation}
where $\text{sim}(\cdot)$ denotes cosine similarity and $\tau$ is a temperature hyperparameter. This loss encourages the model to learn node representations that are invariant to the applied structural perturbations.

\paragraph{Global Structural Alignment via Consistency Constraint}
The InfoNCE loss aligns individual node embeddings but does not explicitly preserve the global similarity structure among nodes. To ensure that the perturbed view retains the overall topological patterns of the original graph, we introduce an additional consistency constraint \cite{wang2020gcn}. Specifically, we first normalize the final-layer embeddings ($L$-th layer) of both views:
\begin{equation}
Z_{\text{raw}} = \begin{bmatrix}Z_{\text{raw}}^{(1)}\\ \vdots \\ Z_{\text{raw}}^{(N)}\end{bmatrix}, \quad Z_{\text{pert}} = \begin{bmatrix}Z_{\text{pert}}^{(1)}\\ \vdots \\ Z_{\text{pert}}^{(N)}\end{bmatrix},
\end{equation}
where
\begin{equation*}
Z_{\text{raw}}^{(u)} = \frac{h_{(u,L)}'}{\|h_{(u,L)}'\|_{2}}, \quad
Z{\text{pert}}^{(u)} = \frac{h_{(u,L)}}{\|h_{(u,L)}\|_{2}}, \quad u=1,\cdots,N
\end{equation*}
and $\|\cdot\|_{2}$ denotes the $L_{2}$-norm of vectors. We then compute the node similarity matrices for both views as:
\begin{equation}
S_{\text{raw}} = Z_{\text{raw}} Z_{\text{raw}}^\top, \quad S_{\text{pert}} = Z_{\text{pert}} Z_{\text{pert}}^\top.
\end{equation}
These matrices encode the global pairwise similarity structure within each view. To enforce consistency, we minimize the Frobenius norm of their difference:
\begin{equation}
\mathcal{L}_{c} = \|S{\text{raw}} - S_{\text{pert}}\|_F^2.
\end{equation}
This constraint ensures that the perturbed view does not deviate arbitrarily from the original graph's global relational pattern, thereby preserving structural integrity.

\paragraph{Joint Optimization Objective}
The overall loss function for SPGCL combines the node-level contrastive loss and the global consistency constraint:
\begin{equation}
\mathcal{L}_{\text{total}} = \mathcal{L}_{\text{task}} + \beta \cdot \mathcal{L}_{\text{I}} + \gamma \cdot \mathcal{L}_{c},
\end{equation}
where $\mathcal{L}_{\text{task}}$ is the supervised classification loss (e.g., cross-entropy) on the labeled nodes, and $\beta, \gamma > 0$ are hyperparameters balancing the contributions of the contrastive and consistency terms. During training, all components are optimized jointly, enabling the model to learn robust node representations that are both discriminative for the downstream task and invariant to structure-preserving perturbations.

\paragraph{Summary}
The proposed SPGCL framework elegantly combines a principled two-stage perturbation with a multi-level contrastive fusion. The perturbation module ensures that the augmented view is both diverse and structure-aware, while the fusion module aligns the two views at node and global levels to learn robust and consistent representations. This integrated design achieves a favorable balance between simplicity and effectiveness, as evidenced by the extensive experiments in the following section.

\section{Experiments}

\subsection{Experimental Setup}
\subsubsection{Datasets}
We evaluate SPGCL and baseline models on ten real-world datasets spanning various domains and structural properties, with statistics summarized in Table~\ref{tab:dataset}. These include:
\begin{itemize}
\item \textbf{Citation networks} (Cora, Citeseer, CoraML) \cite{kipf2017semi,bojchevski2018graph2gauss}: Nodes represent publications and edges denote citations, exhibiting strong homophily \cite{newman2002assortative}.
\item \textbf{Wikipedia networks} (Chameleon, Squirrel, Wiki) \cite{rozemberczki2021multi,Yang_2023}: Nodes are web pages and edges are hyperlinks; these graphs show more heterophilous relationships.
\item \textbf{Social networks} (BlogCatalog, Flickr) \cite{Yang_2023}: Nodes are users and edges indicate friendships or followships.
\item \textbf{Co-purchase network} (Amazon-Photo) \cite{shchur2019pitfallsgraphneuralnetwork}: Nodes are products and edges indicate frequent co-purchases.
\item \textbf{Actor co-occurrence network} (Actor) \cite{Pei2020Geom-GCN}: Nodes are actors and edges denote co-appearance in films.
\end{itemize}
The datasets cover a wide range of sizes, densities, and homophily levels, providing a comprehensive testbed for evaluating robustness and generalization.

\begin{table*}[htbp]
  \centering
  \caption{Description of datasets.}
  \label{tab:dataset}
  \begin{tabular}{l|cccccc}
      \toprule
      \textbf{Dataset} & \textbf{Nodes}($N$) & \textbf{Edges}($|\mathcal{E}|$) & \textbf{Classes} & \textbf{Features}($d$) &\textbf{Train nodes} & \textbf{Val/Test nodes} \\
      \midrule
      \textbf{Cora}  & 2,708 & 5,429 & 7 & 1,433& 140 & 500/1,000 \\
      \textbf{Citeseer}  & 3,327 & 4,732 & 6 & 3,703& 120 & 500/1,000 \\
      \textbf{CoraML}  & 2,995 & 16,316 & 7 & 2,879 & 1,797 &599/599 \\
      \midrule
      \textbf{Chameleon}  & 2,277 & 36,101 & 5 & 2,325 &1,092 & 729/456 \\
      \textbf{Squirrel}  & 5,201 & 217,073 & 5 & 2,089 &2,496 & 1,644/1,041 \\
      \textbf{Wiki}  & 2,405 & 17,981 & 17 & 4,973 &1,443 & 481/481\\
      \midrule
      \textbf{Blog\-Catalog}  & 5,196 & 343,486 & 6 & 8,189 &3,117 & 1,039/1,040  \\
      \textbf{Fl\-ickr}  & 7,575 & 479,476 & 9 & 12,047 &4,545 & 1,515/1,515 \\
      \midrule
      \textbf{Amazo\-n-Photo}  & 7,650 & 238,162 & 8 & 745 &4,590 & 1,530/1,530  \\
      \midrule
      \textbf{Ac\-tor}  & 7,600 & 30,019 & 5 & 932 &3,648 & 2,432/1,520  \\
      \bottomrule
  \end{tabular}
\end{table*}

\subsubsection{Baselines}
We compare SPGCL with three groups of representative methods:
\begin{itemize}
\item \textbf{Spectral GNNs}: GCN \cite{kipf2017semi} and SGC \cite{wu2019simplifying}.
\item \textbf{Spatial GNNs}: GraphSAGE \cite{hamilton2017inductive} and GAT \cite{velickovic2018graph}.
\item \textbf{Graph structure learning methods}: LDS \cite{franceschi2019learning}, Pro-GNN \cite{jin2020graph}, and Geom-GCN \cite{Pei2020Geom-GCN}.
\item \textbf{Perturbation-based regularizer}: DropEdge \cite{rong2020dropedge} applied to GCN.
\end{itemize}

\subsubsection{Implementation Details}
All experiments are conducted on a Linux server with an NVIDIA GeForce RTX 3090 GPU and an AMD EPYC 7282 CPU. We implement SPGCL in Python 3.12.9 using PyTorch 2.5.1. For fairness, we use the same GCN backbone for all GNN-based methods and follow the same semi-supervised splitting strategy as in \cite{fey2019fast}. Hyperparameters are tuned via validation set performance. We report the average accuracy over $5$ independent runs.

\subsection{Node Classification Performance}

\subsubsection{Comparison with Baselines and Perturbation Strategies}
The overall node classification accuracy across all datasets is presented in Table~\ref{tab:node_classification}. We also include a comparison against two naive perturbation strategies—\textbf{Node perturbation} (adding Gaussian noise to features) and \textbf{Edge perturbation} (random edge addition/dropping)—in Table~\ref{tab:results of perturbation comparison} to isolate the benefit of our structured design.

\textbf{Key observations are as follows:}
\begin{itemize}
\item \textbf{Traditional GNNs} (GCN, SGC, GraphSAGE, GAT) perform well on homophilous graphs (Cora, Citeseer) but degrade significantly on heterophilous ones (Chameleon, Squirrel, Actor), highlighting their sensitivity to structural mismatch.
\item \textbf{Graph structure learning methods} (LDS, Pro-GNN) show improved robustness on some heterophilous datasets (e.g., Pro-GNN on Chameleon), but their performance is inconsistent across different graph types.
\item \textbf{DropEdge} helps alleviate over-smoothing on homophilous graphs and achieves competitive results on BlogCatalog and Flickr, but still fails to generalize well under complex structural heterophily.
\item \textbf{SPGCL consistently achieves the best or second-best performance on nearly all datasets}. It exhibits particularly large gains on challenging heterophilous graphs (e.g., +5.6\% over Pro-GNN on Chameleon, +3.6\% on Squirrel) while maintaining strong accuracy on homophilous ones. This demonstrates the broad effectiveness of our structured perturbation.
\item \textbf{SPGCL significantly outperforms naive perturbation strategies} (Table~\ref{tab:results of perturbation comparison}), with an average improvement of 10-15\%. This confirms that the benefits stem from our principled, structure-aware design rather than from random noise injection.
\end{itemize}

These results validate that SPGCL's \emph{coupled stochastic-SVD perturbation}, built upon the generalized perturbation framework \cite{deng2026gadpn}, effectively preserves important global structural patterns while introducing beneficial local diversity. The consistent gains across both homophilous and heterophilous graphs demonstrate that our method successfully bridges the long-standing gap between random and spectral augmentations, leading to more robust and discriminative node representations.

\begin{table*}[htbp]
  \centering
  \caption{Node classification results(\%). (Bold:best; Underline:runner-up.)}
  \label{tab:node_classification}
  \begin{tabular}{l|ccc|ccc|cc|c|c}
    \toprule
    \text{Method} & \text{Cora} & \text{Citeseer} & \text{CoraML} & \text{Chameleon} &\text{Squirrel} & \text{Wiki} &\text{Blog\-Catalog} 
    &\text{Fl\-ickr} & \text{Amazo\-n-Photo} &\text{Ac\-tor}
   \\
    \midrule
    \text{GCN}  & 80.10 & 70.20 & 80.13 &35.21 & 27.74 & 53.66 & 62.75 & 59.54 & 55.23 & 28.91\\
    \text{SGC}  & 80.00 & 71.40 & 32.05 &34.64 & 27.87 & 61.12 & 42.88 & 16.37 & 25.36 & 26.66\\
    \text{GraphSAGE}  & 80.90 & \underline{71.50} &  \underline{89.98} &47.34 & 33.74 & 59.88 & 69.42 & \underline{60.94} & \underline{87.58} & \textbf{33.84}\\
    \text{GAT}  & 80.80 & 68.90 & 86.14 &44.12 & 27.28 & 55.30 & 57.88 & 46.83 & 84.97 & 28.54\\         
    \midrule
    \text{LDS} & 80.60 & 71.12 & 71.62 &36.75 & 27.87 & 45.32 & \underline{77.79} & 58.48 & 52.61 & 28.56\\
    \text{Pro-GNN}  & \underline{81.30} & \underline{71.50} & 85.48 & \underline{56.20} & \underline{34.97} & 74.84 & 78.08 & 57.42 & 70.59 & 30.83\\
    \text{Geom-GCN}  & 70.80 & 59.17 & 76.79 &26.07 & 24.37 & 36.01 & 47.79 & 37.69 & 25.32 & 25.31\\
    \midrule
    \text{DropEdge}  & 80.70 & 70.90 & 85.81 &32.42 & 25.27 & \underline{77.34} & 75.96 & \textbf{61.58} & 75.40 & 28.62\\
    \midrule
    \text{Ours}  & \textbf{83.10} & \textbf{74.30} & \textbf{90.32} & \textbf{61.84} & \textbf{38.62} & \textbf{78.69} & \textbf{79.62} & 58.04 & \textbf{90.84} & \underline{30.92}\\
    \bottomrule
  \end{tabular}
\end{table*}

 \begin{table}[htbp]
  \centering
  \caption{Classification results(\%) of our model, node perturbation, and edge perturbation.}
  \label{tab:results of perturbation comparison}
  \begin{tabular}{l|ccc}
    \toprule
    \text{Method} & \text{Node} & \text{Edge} & \text{Ours}  \\
    \midrule 
    \textbf{Cora}  & 68.10 & 75.40 & \textbf{83.10} \\
    \textbf{Citeseer}  & 64.70 & 60.30 & \textbf{74.30}  \\
    \textbf{CoraML}  & 85.00 & 85.81 & \textbf{90.32}  \\
    \midrule
    \textbf{Chameleon}  & 32.12 & 44.75 & \textbf{61.84}  \\
    \textbf{Squirrel}  & 25.47 & 28.81 & \textbf{38.62}  \\
    \textbf{Wiki}  & 64.97 & 57.08 & \textbf{78.69} \\
    \midrule
    \textbf{Blog\-Catalog}  & 62.25 & 66.80 & \textbf{79.62} \\
    \textbf{Fl\-ickr}  & 32.35 & 32.00 & \textbf{58.04} \\
    \midrule
    \textbf{Amazo\-n-Photo}  & 82.21 & 82.29 & \textbf{90.84}  \\
    \midrule
    \textbf{Ac\-tor}  & 24.32 & 24.75 & \textbf{30.92}  \\
    
    \bottomrule
  \end{tabular}
\end{table}

\subsection{Ablation Studies}
\subsubsection{Ablation on the Perturbation Module}
To validate the design of our perturbation module, we compare SPGCL with three variants:
\begin{itemize}
\item \textbf{SVD}: Use only the SVD-based reconstruction without stochastic removal.
\item \textbf{Edge}: Use only random edge dropping without SVD recovery.
\item \textbf{SVD \& Edge}: Apply SVD first, then perform random edge dropping on the reconstructed view.
\end{itemize}
Results in Table~\ref{tab:results of Ablation} lead to several insights:
\begin{itemize}
\item \textbf{SPGCL outperforms all variants}, demonstrating the necessity of \emph{coupling} stochastic and spectral perturbations. Neither component alone is sufficient.
\item \textbf{Edge alone} often degrades performance, especially on heterophilous graphs, confirming that structure-agnostic random dropping is harmful.
\item \textbf{SVD alone} performs reasonably well, particularly on datasets like CoraML and Wiki, showing the value of global structural awareness.
\item \textbf{SVD \& Edge performs poorly} across all datasets. This is because dropping edges \emph{after} SVD reconstruction randomly destroys the globally consistent patterns recovered by SVD, resulting in a view that is neither diverse nor structure-aware, highlighting the importance of our specific integration order.
\end{itemize}

These findings underscore that our two-stage design-where stochastic removal provides diversity and SVD-guided recovery, grounded in the perturbation theory of \cite{deng2026gadpn}, preserves and enhances structural integrity-is crucial for generating effective contrastive views. The poor performance of the “SVD \& Edge” variant further confirms that simply cascading the two operations fails; the principled integration enabled by the perturbation framework is key.

 \begin{table}[htbp]
  \centering
  \caption{Classification results(\%) of our model and its variants.}
  \label{tab:results of Ablation}
  \begin{tabular}{l|cccc}
    \toprule
    \text{Method} & \text{SVD} & \text{Edge} & \text{SVD\&Edge} & \text{Ours}  \\
    \midrule 
    \textbf{Cora}  & 81.30 & 75.40 &71.00 &\textbf{83.10} \\
    \textbf{Citeseer}  & 67.00 & 60.30&57.60 & \textbf{74.30}  \\
    \textbf{CoraML}  & 89.51 & 85.81 &71.62 &\textbf{90.32}  \\
    \midrule
    \textbf{Chameleon}  & 60.73 & 44.75 &49.91& \textbf{61.84}  \\
    \textbf{Squirrel}  &32.47 & 28.81 &25.79 &\textbf{38.62}  \\
    \textbf{Wiki}  & 72.77 & 57.08 & 60.08&\textbf{78.69} \\
    \midrule
    \textbf{Blog\-Catalog}  & 61.54 & 66.80&38.75 & \textbf{79.62} \\
    \textbf{Fl\-ickr}  & 42.11 & 32.00 & 20.40&\textbf{58.04} \\
    \midrule
    \textbf{Amazo\-n-Photo}  & 88.23 & 82.29 &47.52& \textbf{90.84}  \\
    \midrule
    \textbf{Ac\-tor}  & 24.94 & 24.75 & 22.41&\textbf{30.92}  \\
    
    \bottomrule
  \end{tabular}
\end{table} 

\subsubsection{Ablation on the Global Consistency Loss $\mathcal{L}_c$}
To evaluate the contribution of the proposed global similarity-matrix consistency regularizer, we remove $\mathcal{L}_c$ from the objective while keeping all other components unchanged. As shown in Table~\ref{tab:ablation_lc}, discarding $\mathcal{L}_c$ consistently leads to a performance drop on both homophilous (Cora, Citeseer) and heterophilous (Chameleon, Squirrel) datasets. This indicates that $\mathcal{L}_c$ provides complementary supervision beyond the node-level InfoNCE alignment by explicitly stabilizing the global relational structure across the two views, which is particularly beneficial for maintaining structural consistency under perturbation.

\begin{table}[htbp]
    \centering
    \caption{Ablation study on the global consistency loss $\mathcal{L}_c$ (Acc.\%).}
    \label{tab:ablation_lc}
    \begin{tabular}{l|ccc|ccc}
        \toprule
        \textbf{Variant} & \textbf{Cora} & \textbf{Citeseer}  & \textbf{Chameleon} & \textbf{Squirrel}\\
        \midrule
         $+\mathcal{L}_c$& \textbf{83.10} & \textbf{74.30} & \textbf{61.84} & \textbf{38.62}  \\
        $-\mathcal{L}_c$ & 82.20 & 72.40 & 61.62 & 37.66  \\
        \bottomrule
    \end{tabular}
\end{table}

\subsection{Robustness Analysis under Structural Noise}
To evaluate robustness under increasing structural noise, we simulate corrupted graphs by randomly removing 20\% and 30\% of edges, and compare SPGCL with baselines. Results in Table~\ref{tab:robustness} show:
\begin{itemize}
\item Most baselines (GCN, SGC, GraphSAGE, GAT) suffer noticeable accuracy drops as the perturbation ratio increases.
\item SPGCL maintains the highest or competitive accuracy under both corruption levels, with a \textbf{smaller relative decline}. For example, on Cora, SPGCL drops by only 6.4\% when 30\% of edges are removed, whereas GCN drops by 9.7\%.
\end{itemize}

This resilience stems from the \emph{theoretically grounded recovery mechanism} provided by the generalized perturbation framework \cite{deng2026gadpn}, which can reconstruct important edges even under significant random deletion, coupled with the contrastive fusion that aligns representations across noisy views. This demonstrates SPGCL's practical value in noisy real-world scenarios and further validates the utility of the underlying perturbation theory beyond its original domain.

\begin{table}[htbp]
  \centering
  \caption{Comparison between competing methods and our model under the random perturbation scenarios.}
  \label{tab:robustness}
  \renewcommand{\arraystretch}{1.15}
  \setlength{\tabcolsep}{5pt}
  \begin{tabular}{l|cc|cc|cc}
    \toprule
    \textbf{Method} &
    \multicolumn{2}{c|}{\textbf{Cora}} &
    \multicolumn{2}{c|}{\textbf{CiteSeer}} &
    \multicolumn{2}{c}{\textbf{Cora\-ML}} \\
    & 0.2 & 0.3 & 0.2 & 0.3 & 0.2 & 0.3 \\
    \midrule
    GCN        & 76.13 & 72.30 & 64.30 & 65.80 & 77.30 & 73.62 \\
    SGC        & 76.10 & 72.70 & \textbf{67.80} & \textbf{65.90} & 81.97 & 78.63 \\
    SAGE       & 74.80 & 71.70 & 67.10 & 64.00 & 85.98 & \textbf{82.47} \\
    GAT        & 74.80 & 71.70 & 66.30 & 63.50 & 79.97 & 73.62 \\
    \midrule
    LDS        & 74.90 & 72.00 & 67.30 & 64.10 & 59.43 & 52.25 \\
    Gemo       & 66.60 & 68.90 & 49.60 & 58.10 & 74.79 & 74.63 \\
    \midrule
    DropEdge   & 75.70 & 72.50 & 65.90 & 63.30 & 76.13 & 72.45 \\
    \midrule
    Our        & \textbf{80.50} & \textbf{74.14} & 66.20 & 64.40 & \textbf{88.98} & 76.74 \\
    \bottomrule
  \end{tabular}
\end{table}

\subsection{Parameter Sensitivity Analysis}
\label{subsec:analysis_pq}
\begin{figure}[t]
  \centering
  \begin{subfigure}[t]{0.47\linewidth}
    \centering
    \includegraphics[width=\linewidth]{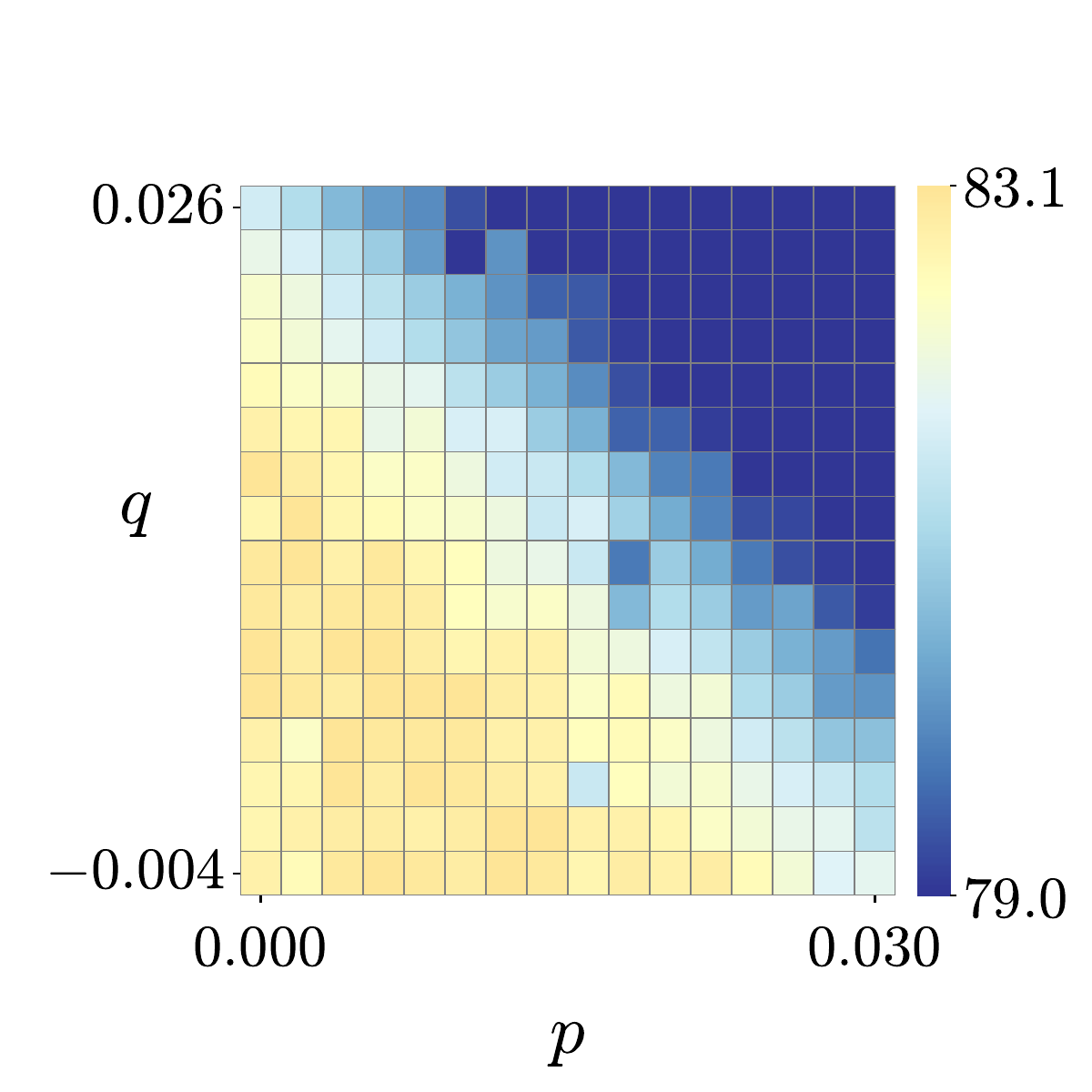}
    \caption{Cora}
    \label{fig:pq_cora}
  \end{subfigure}\hspace{0.02\linewidth}
  \begin{subfigure}[t]{0.47\linewidth}
    \centering
    \includegraphics[width=\linewidth]{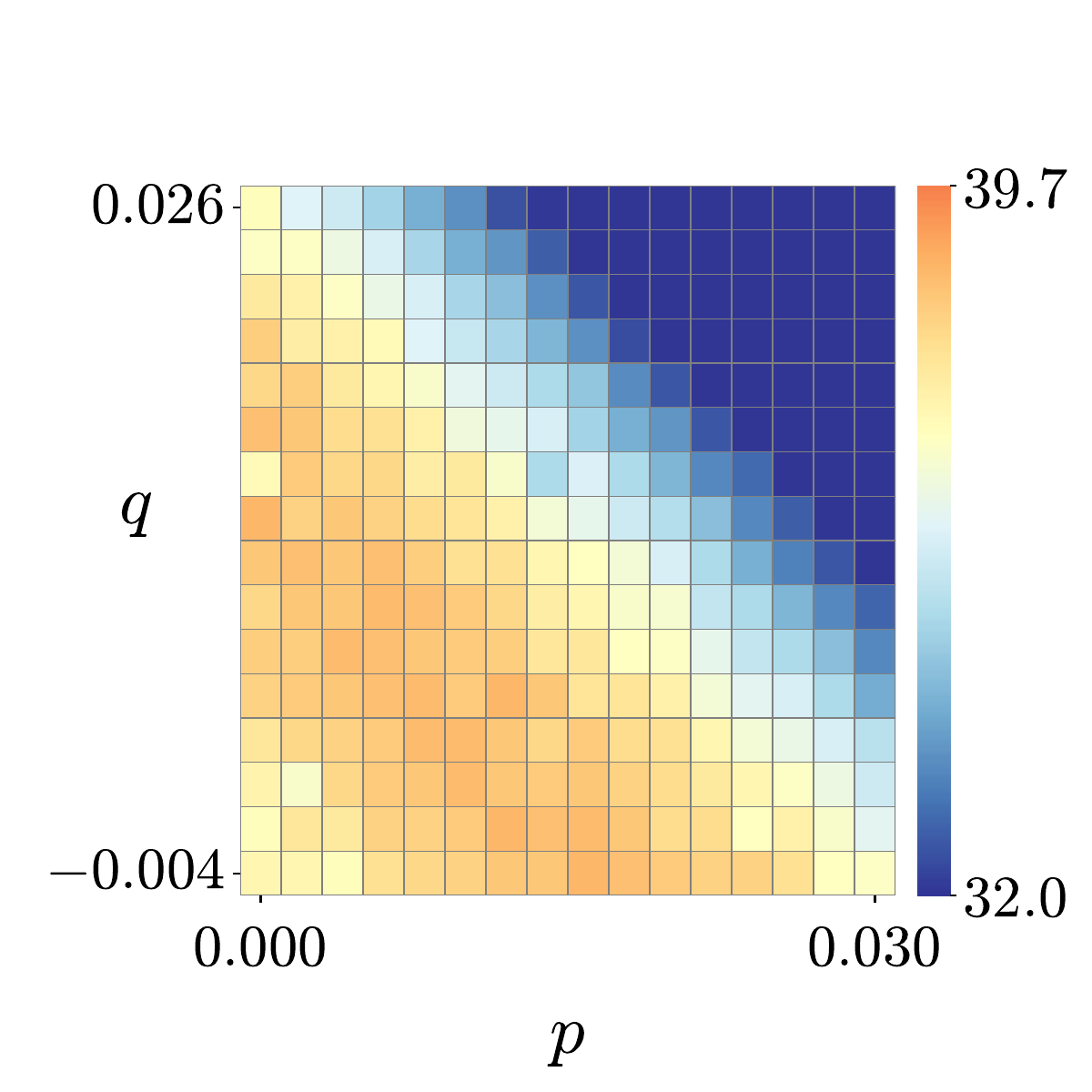}
    \caption{Squirrel}
    \label{fig:pq_squirrel}
  \end{subfigure}
  \caption{Sensitivity analysis of $p$ and $q$.}
  \label{fig:pq_analysis}
\end{figure}

We analyze the sensitivity of SPGCL to two key hyperparameters: the base edge drop ratio $p$ and the offset $q$ in Eq.~\eqref{eq:P_pq}, which jointly control the scale of perturbation ($P$). We perform a 2D grid search over $p \in [0.000, 0.030]$ and $q \in [-0.004, 0.026]$ and report the resulting accuracy on Cora and Squirrel in Fig.~\ref{fig:pq_analysis}.

\textbf{Observations:} The performance exhibits a clear unimodal trend. When $P$ is too small (under-perturbation), the contrastive view is overly similar to the original, limiting diversity. As $(p+q)$ increases to a moderate range, performance improves and reaches a peak. However, overly large $P$ (over-perturbation) introduces many low-confidence edges, leading to graph densification and performance degradation. This analysis confirms that our method is not overly sensitive to hyperparameters within a reasonable range and provides practical guidance for tuning.

\subsection{Visualization of Learned Embeddings}
\begin{figure*}[htbp]
    \centering
    \begin{subfigure}{0.19\linewidth}
        \centering
        \includegraphics[width=\linewidth]{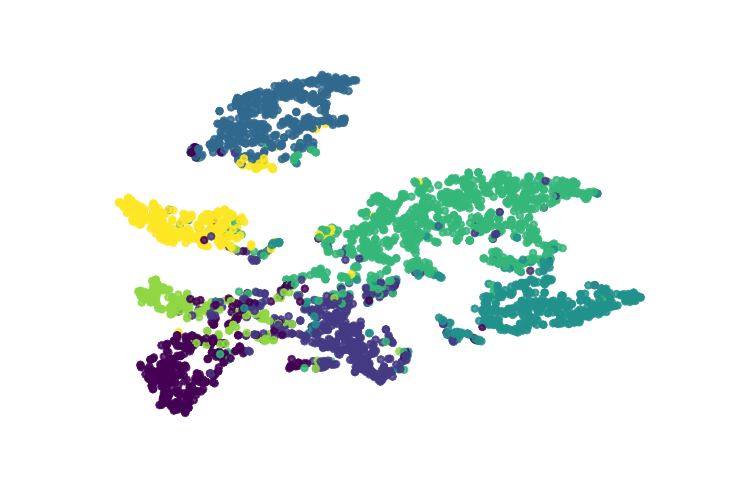}
        \caption{GCN}
    \end{subfigure}%
    \hspace{-0.2em}
    \begin{subfigure}{0.19\linewidth}
        \centering
        \includegraphics[width=\linewidth]{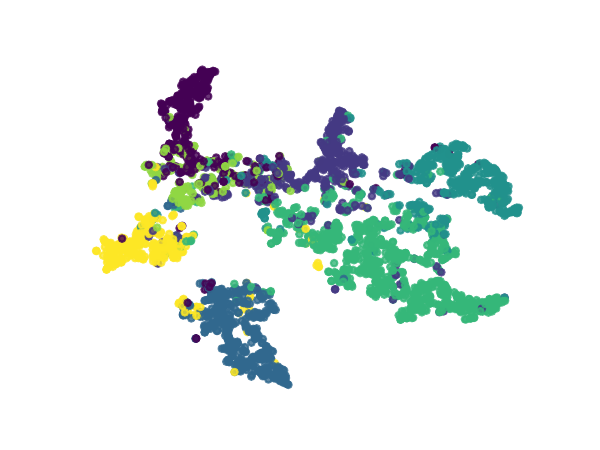}
        \caption{LDS}
    \end{subfigure}%
    \hspace{-0.2em}%
    \begin{subfigure}{0.19\linewidth}
        \centering
        \includegraphics[width=\linewidth]{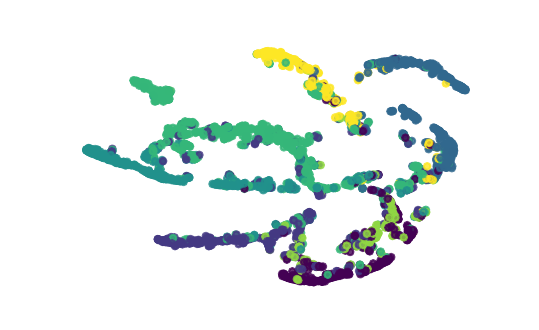}
        \caption{Geom-GCN}
    \end{subfigure}%
    \hspace{-0.2em}%
    \begin{subfigure}{0.19\linewidth}
        \centering
        \includegraphics[width=\linewidth]{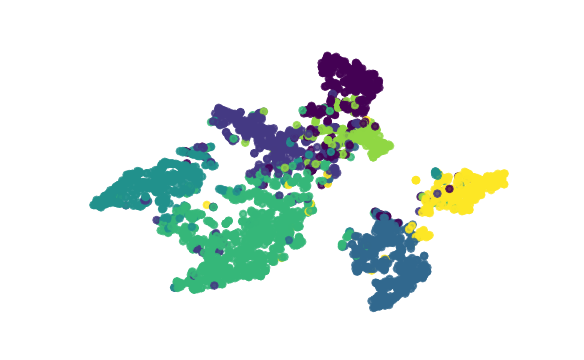}
        \caption{DropEdge} 
    \end{subfigure}%
    \hspace{-0.2em}%
    \begin{subfigure}{0.19\linewidth}
        \centering
        \includegraphics[width=\linewidth]{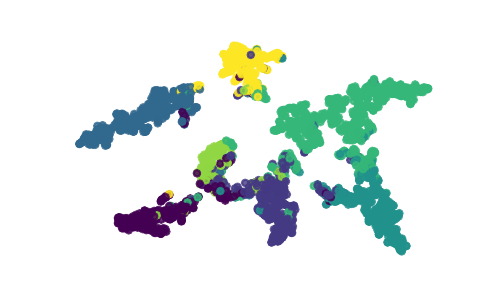}
        \caption{Ours}
    \end{subfigure}
    \caption{t-SNE visualization of node embeddings on CoraML. 
    While baselines (a-d) show blurred boundaries, SPGCL (e) exhibits distinct class separation and compact intra-class clustering, 
    demonstrating the effectiveness of the proposed fusion strategy.}
    \label{fig:vis}
\end{figure*}

We visualize the test-set embeddings learned by different models on CoraML using t-SNE \cite{van2008visualizing}. As shown in Fig.~\ref{fig:vis}:
\begin{itemize}
\item Baselines (GCN, LDS, Geom-GCN, DropEdge) produce embeddings with blurred class boundaries and overlapping clusters.
\item SPGCL generates well-separated, compact clusters with clearer margins.
\end{itemize}
This visualization confirms that SPGCL learns more discriminative and structurally coherent representations, aligning with the quantitative performance gains.

\subsection{Summary of Experimental Findings}
Through extensive experiments, we demonstrate that:
\begin{enumerate}
\item \textbf{Superior Performance:} SPGCL consistently outperforms state-of-the-art GNNs and graph structure learning methods across diverse graph types, proving its general effectiveness.
\item \textbf{Design Validity:} The structured perturbation is superior to naive random perturbations, and ablation studies confirm the necessity of both stochastic removal and SVD-guided recovery.
\item \textbf{Enhanced Robustness:} SPGCL exhibits stronger resilience under increasing structural noise, making it suitable for imperfect real-world graphs.
\item \textbf{Beneficial Components:} The global consistency loss $\mathcal{L}_c$ provides complementary structural alignment, and the perturbation scale can be effectively tuned via $p$ and $q$.
\end{enumerate}

These results collectively affirm that SPGCL achieves an excellent balance of \textbf{simplicity} in design and \textbf{power} in performance, offering a robust solution for graph contrastive learning. More importantly, they demonstrate how a principled integration of random and spectral augmentations-enabled by a rigorous perturbation framework-can lead to significant empirical gains across diverse graph types and noise levels.

\section{Conclusion}

This paper proposes \textbf{SPGCL}, a simple yet powerful framework for robust graph contrastive learning via SVD-guided structural perturbation. SPGCL directly addresses the long-standing dichotomy in graph augmentation strategies: random perturbations are structure-agnostic, while purely spectral methods often produce dense and deterministic views. The key to our solution lies in recognizing that the recently developed \emph{generalized singular value perturbation theory} \cite{deng2026gadpn}-which extends classical structural perturbation \cite{lu2015toward} from symmetric to arbitrary graphs-can serve as the missing bridge to connect these two disparate paradigms.

Building upon this theoretical foundation, we design a principled two-stage strategy that first applies lightweight stochastic edge removal to ensure view diversity, and then employs truncated SVD with perturbation-guided scoring to recover and enhance topologically important edges. This concise yet theoretically grounded design delivers three key advantages: (1) \textbf{robustness to accidental edge deletion}, (2) \textbf{enrichment with missing high-order connections}, and (3) \textbf{explicit control over structural discrepancy}. Furthermore, we introduce a contrastive fusion module with a global consistency constraint to align node embeddings across views at both local and global levels, ensuring invariant and structurally coherent representations.

Extensive experiments on ten benchmark datasets demonstrate that SPGCL consistently outperforms state-of-the-art GNNs and graph structure learning methods. It achieves particularly significant gains on challenging heterophilous graphs while maintaining strong performance on homophilous ones. Ablation studies confirm the necessity of both stochastic and spectral components, and robustness analyses show SPGCL's superior stability under increasing structural noise. The visualization of learned embeddings further corroborates that SPGCL produces more discriminative and well-separated clusters.

In summary, SPGCL provides an effective and generalizable solution for learning robust graph representations in noisy and heterogeneous environments. Our work also illustrates the versatility of the generalized SVD perturbation framework, showcasing how a method originally developed for graph structure learning can be creatively adapted to solve a fundamentally different challenge in contrastive learning. Future work may explore several promising directions: (1) adaptive perturbation strategies that can dynamically adjust the augmentation strength (e.g., drop ratio $p$, recovery offset $q$) based on graph properties (e.g., homophily level) or training dynamics, moving beyond static hyper-parameter tuning; (2) extending the paradigm to dynamic graphs (where structure evolves over time) and heterogeneous graphs (with multiple node/edge types), which requires re-thinking structural perturbations in time-aware and relation-aware manners; and (3) investigating the integration of SPGCL with other advanced self-supervised learning paradigms for even broader applicability.

\vspace{-33pt}
\begin{IEEEbiography}[{\includegraphics[width=1in,height=1.25in,clip,keepaspectratio]{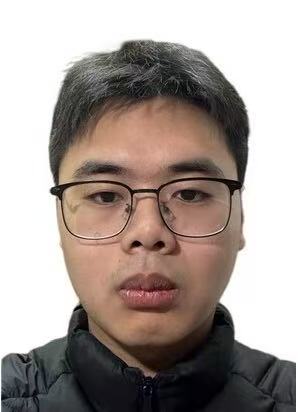}}]{Hao Deng} received the B.S. degree in electronic and information engineering from Southwest Petroleum University, Chengdu, China, in 2024. He is currently pursuing the M.S. degree in control science and engineering with Xidian University, Xi'an, China. His current research interests include graph neural networks and graph structure learning.    
\end{IEEEbiography}
\vspace{-33pt}
\begin{IEEEbiography}[{\includegraphics[width=1in,height=1.25in,clip,keepaspectratio]{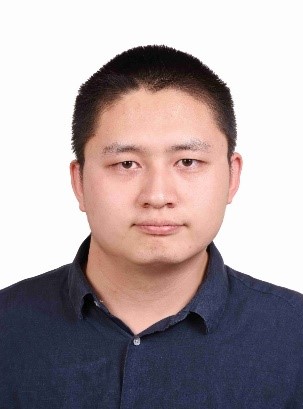}}]{Zhang Guo} (Member, IEEE) received the B.S., M.S., and Ph.D. degrees in electrical engineering from Xi'an Jiaotong University, Xi'an, China, in 2013, 2016 and 2021, respectively. He is a Member of the IEEE. He is currently an Assistant Professor at the School of Artificial Intelligence, Xidian University. His main research interests include spiking neural network, neuromorphic computing and artificial intelligence. 
\end{IEEEbiography}
\vspace{-33pt}
\begin{IEEEbiography}[{\includegraphics[width=1in,height=1.25in,clip,keepaspectratio]{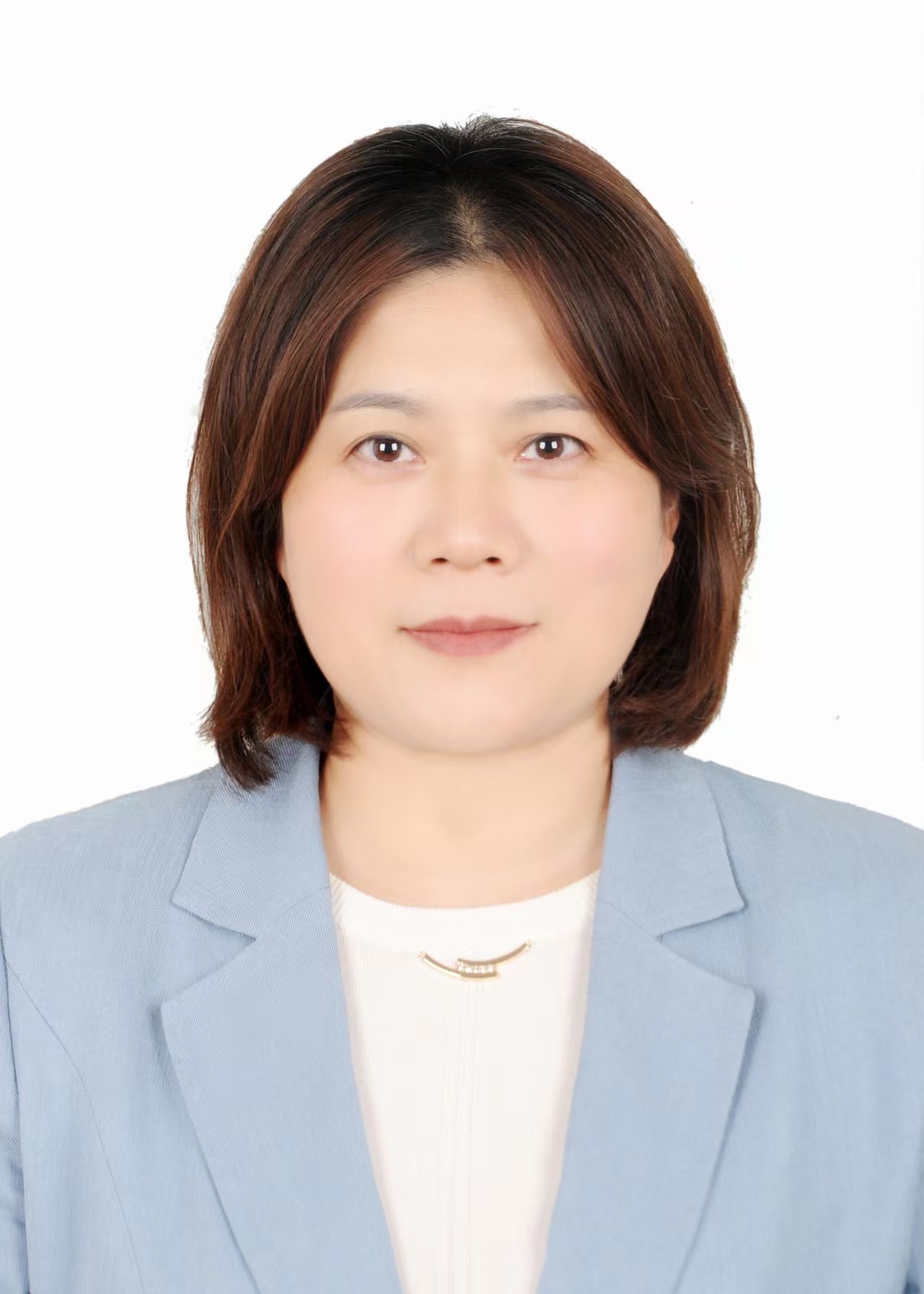}}]{Shuiping Gou} (Senior Member, IEEE) received the B.S. and M.S. degrees in computer science and technology and the Ph.D. degree in pattern recognition and intelligent systems from Xidian University, Xi’an, China, in 2000, 2003, and 2008, respectively. She is currently a Professor with the Key Laboratory of Intelligent Perception and Image  Understanding of Ministry of Education of China, Xidian University. Her research interests include machine learning, data mining, and medical image analysis.
\end{IEEEbiography}
\vspace{-33pt}
\begin{IEEEbiography}[{\includegraphics[width=1in,height=1.25in,clip,keepaspectratio]{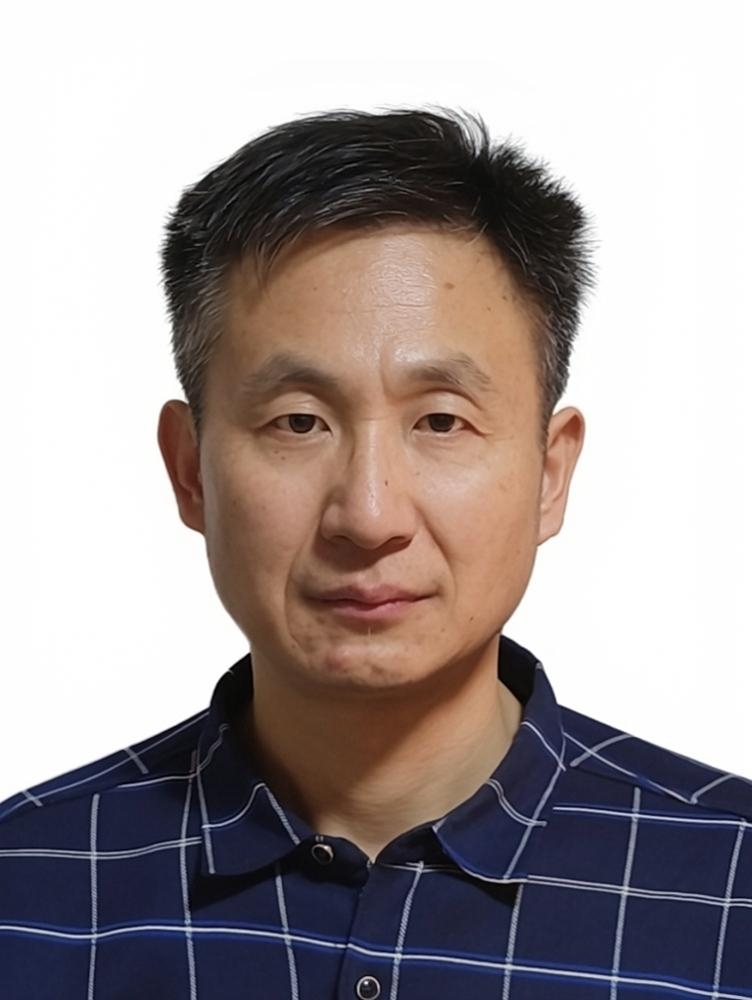}}]{Bo Liu} (Member, IEEE) received the B.A. degree in biotechnology from Zhejiang University, Hangzhou, China, in 2001, and the Ph.D. degree in applied mathematics from Fudan University, Shanghai, China, in 2010. He was a Post-Doctoral Fellow with School of Life Sciences, Fudan University from 2010 to 2012, and a Post-Doctoral Researcher with the Institute of Industrial Science, The University of Tokyo from 2012 to 2014. Since 2014, he has been a professor with Key Laboratory of Intelligent Perception and Image Understanding of the Ministry of Education of China, School of Electronic Engineering (School of Artificial Intelligence since November, 2017), Xidian University, Xi’an. His  research interests are the theory and applications of networked systems, including dynamical complex networks, neural networks and deep learning.
\end{IEEEbiography}

\bibliographystyle{IEEEtran} 
\bibliography{references}     

\end{document}